\documentclass[letterpaper, 10pt, journal]{IEEEtran}  

\IEEEoverridecommandlockouts                              

\usepackage{subfigure}
\usepackage{graphicx}
\usepackage{bbm}
\usepackage{amssymb}
\usepackage{amsmath}
\usepackage{wrapfig}
\usepackage{mathptmx} 
\usepackage{microtype} 
\usepackage{booktabs}
\usepackage{hyperref}
\usepackage{color}
\usepackage{placeins}

\title{\LARGE \bf How Much Do Unstated Problem Constraints\\ Limit Deep Robotic Reinforcement Learning?}

\author{W. Cannon Lewis II, Mark Moll, and Lydia E. Kavraki \thanks{W. Cannon Lewis II, Mark Moll, and Lydia E. Kavraki are with the Department of Computer Science, Rice University, Houston TX, USA\@. \texttt{wcannon@rice.edu}, \texttt{mmoll@rice.edu}, \texttt{kavraki@rice.edu}. This work has been supported in part by NSF 1718478 and Rice University Funds.}%
}


\begin{document}

\maketitle
\thispagestyle{empty}
\pagestyle{empty}

\begin{abstract}
    Deep Reinforcement Learning is a promising paradigm for robotic control
    which has been shown to be capable of learning policies for
    high-dimensional, continuous control of unmodeled systems. However, Robotic
    Reinforcement Learning currently lacks clearly defined benchmark tasks,
    which makes it difficult for researchers to reproduce and compare against
    prior work.  ``Reacher'' tasks, which are fundamental to robotic
    manipulation, are commonly used as benchmarks, but the lack of a formal
    specification elides details that are crucial to replication. In this paper
    we present a novel empirical analysis which shows that the unstated spatial
    constraints in commonly used implementations of Reacher tasks make it
    dramatically easier to learn a successful control policy with Deep
    Deterministic Policy Gradients (DDPG), a state-of-the-art Deep RL
    algorithm.  Our analysis suggests that less constrained Reacher tasks are
    significantly more difficult to learn, and hence that existing de facto
    benchmarks are not representative of the difficulty of general robotic
    manipulation. 
\end{abstract}

\def\abstractname{Note to Practitioners}
\begin{abstract}
    This paper seeks to understand how well common algorithms for learning
    robotic manipulation perform and generalize.  Though Deep Reinforcement
    Learning has been the subject of much attention lately, the reported
    results may be difficult to replicate in practice. In particular, this
    paper shows how much the minutiae of a robotic problem definition can
    impact learning. We find that for a particular sort of robotic task, a
    popular Deep Reinforcement Learning algorithm only succeeds when
    non-physical limitations are placed on the problem. This suggests that care
    must be taken when applying these sorts of learning methods to robotic
    problems in the real world, as previously reported results may fail to
    generalize in surprising ways. We also seek to give some intuition as to
    why certain problem limitations may make learning easier. This paper may be
    of interest to anyone considering applying Reinforcement Learning for a
    real-world robotic problem, especially during the problem formulation
    phase. 
\end{abstract}


\section{Introduction}
\label{sec:intro}
Robotic autonomy is challenging for several reasons: robots have many
degrees of freedom, require an agent capable of continuous observation and
action, and can exhibit action and sensing uncertainty. For many problems such
as manipulation under uncertainty, navigation in unknown environments, or
interaction with human beings, it is difficult or impossible to model the
environment well a priori.  It is for this reason that machine learning
methods, which adapt to new environments and tasks, are a promising frontier in
robotic autonomy.  Reinforcement Learning (RL) \cite{sutton1998reinforcement}
is a particularly promising paradigm, as defining an RL problem requires only
the specification of a reward function encoding success. 

In recent years, Deep RL (which extends RL with deep neural networks) has had
demonstrable success on manipulation tasks
\cite{gu2017deep,levine2016end,vevcerik2017leveraging,andrychowicz2017hindsight}.
In reaction to these successes there has been a push toward standardization of
benchmarks and testing conditions used to evaluate Deep RL methods
\cite{duan2016benchmarking,henderson2017deep}. Simulation suites such as OpenAI
Gym \cite{brockman2016openai} and MuJoCo \cite{todorov2012mujoco} have enabled this standardization, but no existing work has
shown that the de facto benchmark tasks are truly representative of the
challenges presented by autonomous motion of commercially available robot
manipulators. More work is needed to clarify precisely when Deep RL
methods work well for general robotic autonomy.

\begin{figure}[bt]
    \centering
    \includegraphics[width=\linewidth]{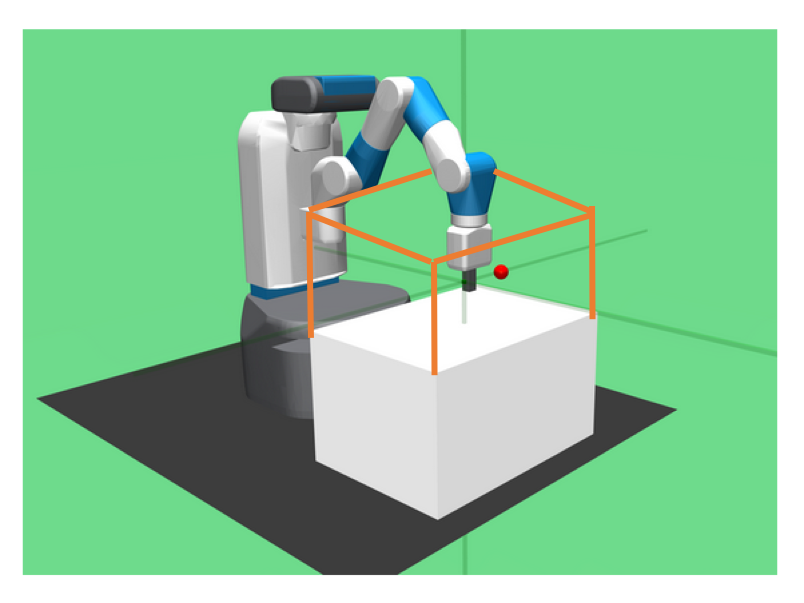}
    \vspace*{-\baselineskip}
    \caption{
      \label{fig:constrained_table}An example of
    a typical goal constraint region in the `FetchReach-v0' environment, a
    commonly-used example of the ``Reacher'' tasks that we analyze in this
    paper. The orange box here is approximately the constraint region from
    which goals are sampled. Image from \cite{plappert2018multi} with added
    constraint visualization.}
\end{figure}

More recently, the authors of \cite{henderson2017deep} have shown that several
axes of variation serve as confounding variables in reported Deep RL results.
They demonstrate that hyperparameter setting, network architecture, reward
scaling, random seeds, environment specification, and codebase choice can have
significant impact on empirical learning behavior. Our goal in this paper is to
extend the binary observation that environment specification can affect
learning to a qualitative one; we wish to understand \emph{to what degree}
small changes in problem description impact the difficulty of a particular
family of robotic tasks. We show this by introducing subtle variations of a
popular robotic RL environment, the ``Reacher'' family of tasks. Indeed,
\cite{henderson2017deep} performs some of their analysis on the ``Reacher''
task variant introduced by \cite{plappert2018multi}, and so our analysis is
highly related to this prior work. ``Reacher'' tasks are a prime example of the
fundamental manipulation and locomotion tasks found in prior work
\cite{Kober2013,kohl2004policy,plappert2018multi}. These tasks are necessary
building blocks to more complex control tasks such as search-and-rescue or
construction, but there has been no extensive analysis of how their
specification can affect learning.

The main contribution of this paper is an analysis showing that the ``Reacher''
tasks used in prior works are not representative of the full difficulty
presented by general robotic manipulation (or even basic pick-and-place tasks).
More precisely, we show that
existing benchmarks (e.g., those proposed in \cite{plappert2018multi})
constrain goal sampling, which has a significant impact on learning. We perform
this analysis by constructing a series of ``Reacher'' tasks which interpolate
between tasks similar to prior work and a more general unconstrained task.
``Reacher'' tasks challenge an agent to use low-level (joint position, velocity, or
torque) control to move the end effector of a manipulator to a point
in the robot's workspace. Note that if position control is used, this family of tasks
amounts to learning the inverse kinematics of the manipulator; if velocity or
torque control is used then the task is closely related to learning inverse
dynamics \cite{craig2005introduction}. Thus, ``Reacher'' tasks expose some
fundamental challenges in robotics. However, prior work uses goal constraint
regions to restrict the effective workspace of the manipulator being controlled
to such a degree that the underlying learning problem is changed and, we argue,
simplified. Our empirical analysis using the state-of-the art DDPG
\cite{lillicrap2015continuous} algorithm on a simulated UR5 robot supports this
point and shows that this task comparison is apt, as we find that the DDPG
fails to generalize in unexpected ways as the effective workspace is expanded.
Following the methods used by \cite{henderson2017deep}, we fix our algorithm,
code, and hyperparameter settings across all experiments, and focus our
analysis on the task definition. 

This analysis is supported and systematized by a software framework (ROSGym)
that we developed to connect standard implementations of RL algorithms to
commonly used robot control software. More precisely, we wrote a Python
interface that integrates the Robot Operating System (ROS)
\cite{quigley2009ros} and OpenAI Gym, which we then used to generate the
results in this paper. The flexibility of ROS allowed
us to easily compare variations of the ``Reacher'' task specification in order
to better understand the influence of factors such as the number of joints and
goal constraint region on learning.

\section{On Deep RL}
In this work, we focus our analysis on the specification of ``Reacher''
benchmark tasks for Robotic RL. In order to eliminate the other sources of
variability identified by \cite{henderson2017deep}, we use a fixed learning
algorithm (DDPG) and fixed hyperparameters to perform our experiments. In this
section, we will provide the minimal necessary background in Deep RL to
contextualize our choice of DDPG for this analysis.

\subsection{RL Background}
\label{sec:background} 
We consider a standard reinforcement learning problem definition
\cite{sutton1998reinforcement}, in which we model a task of interest as a
learning agent interacting with a Markov Decision Process (MDP). An MDP $M$ is
a tuple $M = (S, S_0, A, P, R)$ describing an environment which an agent
interacts with in discrete time steps $t \in \{0\ldots T\}$. More precisely, at each
time step $t$ our agent occupies a state $s_t \in S$, initially sampled from
$S_0$. At each time step, the agent takes an action $a_t \in A$, and
experiences a probabilistic state transition according to the probability
distribution over $s_{t+1}$ defined by the transition function $s_{t+1} \sim
P(s_t, a_t, s_{t+1}) = Pr(s_{t+1} | s_t, a_t)$.  As a result of this action,
the agent also receives a reward $r_t = R(s_t, a_t)$. The tuple $(s_t, a_t,
s_{t+1}, r_t)$ is typically called a \emph{transition}, and the full sequence
of these transitions over $t \in \{0 \ldots T\}$ is called a \emph{trajectory}
or \emph{rollout}.  In this paper we restrict our attention to continuous
control; specifically, the case where $S = \mathbb{R}^m$ and $A = \mathbb{R}^n$ for $m, n \in
\mathbb{N}$.

In reinforcement learning we are concerned with learning a policy $\pi : S
\rightarrow A$ which maximizes the total reward $R_T = \Sigma_{t=0}^T r_t =
\Sigma_{t=1}^T R(s_t, \pi(s_t))$. This policy may also be probabilistic, in
which case we learn $\pi : S \rightarrow \mathbb{P}[A]$, where $\mathbb{P}[A]$
is the set of probability distributions over $A$, and seek to maximize
$\mathbb{E}_{s_0 \sim S_0, a_t \sim \pi(s_t), s_{t+1} \sim P(s_t, a_t, s_{t+1})} [R_T]$.
In practice, RL methods often modify this definition of the total reward to
include a discount for future states. This discounted total reward is called the
value function, and is defined as $V_{\pi}(s) = \mathbb{E}[\Sigma_{i = t}^T
\gamma^{(i - t)} R(s_i, a_i) | s_t = s]$, where $a_t \sim \pi(s_t)$ at all time
steps. This definition of the value function naturally gives rise to the
action-value function $Q_{\pi}(s, a) = \mathbb{E}_{s_{t+1} \sim
P}[V_{\pi}(s_{t+1}) | s_t = s, a_t = a]$, which intuitively assigns a value to
being in a state $s_t$ and taking action $a_t$, while prioritizing earlier
sources of reward.

From previous results \cite{sutton1998reinforcement}, it is known that the
action-value function satisfies the Bellman equation:
\begin{equation*}
    Q_{\pi}(s, a) = \mathbb{E}_{s' \sim P} [R(s, a) + \gamma
    \mathbb{E}_{a' \sim \pi} [Q_{\pi}(s', a')]].
\end{equation*}
The recursive nature of this equation allows RL methods to iteratively estimate
the action-value function from experience. If $Q_{\pi}$ is represented by a
function approximator with parameters $\theta_Q$, we can derive a
differentiable loss:
\begin{equation}
    \label{eq:loss}
    L(\theta_Q) = \mathbb{E}_{s_t \sim P, a_t \sim \pi}[(Q_{\pi}(s_t, a_t |
    \theta_Q) - y_t)^2],
\end{equation}
where
\begin{equation*}
    y_t = R(s_t, a_t) + \gamma Q_{\pi}(s_{t+1}, \pi(s_{t+1}) | \theta_Q).
\end{equation*}
Gradients from this loss can then be used to adjust the parameters $\theta_Q$
and improve the approximation of $Q_{\pi}$ \cite{mnih2015human}.

In problems with finite action spaces, performing the above optimization and
using the greedy policy which selects the action with the highest action-value
at every time step is known as \emph{Q-learning}. However, for problems with continuous
action spaces it is impractical to find the optimal action at each
time step. In \cite{lillicrap2015continuous} the authors show that this
problem can be made tractable using an actor-critic method in which both the
deterministic policy $\mu$ and action-value function $Q$ are approximated by
deep neural networks, which they term Deep Deterministic Policy Gradients
(DDPG). By decomposing \eqref{eq:loss} and isolating the policy
component of the action-value function gradient, DDPG allows an agent to
optimize a policy over a continuous action space. Optimizing a policy in this
way causes learning instability, and so DDPG utilizes an \emph{experience
replay buffer} to decorrelate experienced transitions and improve stability.
Transitions are added to this experience replay buffer when the agent receives
them from the environment, and then the agent samples training examples from
the buffer during training. We refer the interested reader to
\cite{lillicrap2015continuous} for further algorithmic details.

\subsection{Why DDPG?}
\label{sec:relwork}
In this paper, we focus on Deep RL as applied to robotic control, particularly
in manipulation settings. Among recent Deep RL methods, DDPG demonstrates a
number of desirable features.  First, DDPG learns continuous control policies
which eliminate the need for action discretization, the previously dominant
methodology enabling robotic RL \cite{Kober2013}.  Second, DDPG learns a
deterministic control policy, which is advantageous for robotic applications
because the learned policy can be reproducibly tested and verified once
learning has converged. Though other methods such as TRPO
\cite{schulman2015trust} and Soft Q-Learning \cite{haarnoja2017reinforcement}
are promising for problems with continuous action spaces, these methods learn
stochastic policies which are more difficult to verify.  Third, DDPG is
model-free, which means that it can be applied to novel tasks and robots
without extensive feature engineering or incorporation of expert knowledge.
Finally, because DDPG is an off-policy method, it can be modified to
make use of additional sources of experience (such as human
demonstration) which have been shown to improve
learning~\cite{vevcerik2017leveraging}.

To the best of our knowledge, few results exist which successfully apply Deep
RL on simulated or real commercially available robots. Some of the most
impressive results in this field use demonstration for initialization or make
use of a dynamics model to simplify the learning task
\cite{vevcerik2017leveraging,peng2018deepmimic,gu2016continuous}. Model-free
methods are often demonstrated on a variety of simulated tasks from MuJoCo or
OpenAI Gym, and occasionally in tabletop manipulation tasks on a commercially
available robot such as a Fetch, UR5, or Baxter
\cite{gu2017deep,rusu2016sim,andrychowicz2017hindsight}. However, examining the
robotic benchmarks proposed in \cite{plappert2018multi} or implemented in
MuJoCo reveals a core similarity with manipulation benchmarks commonly
demonstrated on commercial robots: task goals (such as goal end effector
position) are sampled from a \emph{goal constraint region} above a ``table'' surface.
Figure~\ref{fig:constrained_table} shows a visualization of a typical goal
constraint region. Recently, \cite{NIPS2017_7233} and \cite{henderson2017deep}
have previously argued that seemingly innocuous unstated assumptions such as
algorithm implementation, parameterization, initialization, and reward scale
can have inordinate effects on the success of learning. Here, we argue that the
goal constraint region is another significant assumption which affects robotic
reinforcement learning. This goal constraint region is an often unstated part
of the Reacher specification, and our recognition of this phenomenon comes from
examining the publicly available implementations of environments in
\cite{plappert2018multi,todorov2012mujoco}\footnote{See, e.g.,
\url{https://github.com/openai/gym} and
\url{https://github.com/openai/mujoco-py}.}.

\section{Methods}
\label{sec:methods}
\subsection{Algorithmic Details} In this work, we conduct experiments using
DDPG \cite{lillicrap2015continuous} with Hindsight Experience Replay (HER)
\cite{andrychowicz2017hindsight}. We implemented our own version of DDPG for
this purpose, which we intend to open-source along with the ROSGym interface
described below.  The authors of \cite{andrychowicz2017hindsight} demonstrate
that, in reinforcement learning tasks similar to the ``Reacher'' tasks
considered in our current work, augmenting the learning agent's experience with
counterfactual experience can speed up learning convergence and result in a
higher success rate for the convergent policy. We employ HER by, for each
episode of training, appending a modified trajectory to DDPG's experience
replay buffer where the rewards $r_t$ are re-calculated as if the final
end effector position reached by the agent during the episode was the goal
position. 

While conducting this research, we examined a number of other modifications to
DDPG and our environment specification that are not employed in the following
experiments. Some prior work (e.g., \cite{schaul2015prioritized}) suggests that
Prioritized Experience Replay may improve learning stability and rate of
convergence, but our experience was that this tended to destabilize or
prevent learning. The experiments below utilize uniform sampling from the
experience replay buffer.  Other work~\cite{andrychowicz2017hindsight} has
suggested that sparse rewards give rise to better learning than dense rewards,
but we found the opposite to be true and so used the dense reward formulated in \eqref{eq:reward}. 

\subsection{Environmental Specification}
When conducting an experiment using RL, an experimenter's choice of $(S, S_0,
A, P, R)$ can have a significant effect on learning.  Though we developed our
own simulated environment using ROS, we took inspiration from the existing
``Reacher-v2'' implementation in OpenAI Gym in order to follow previous
results. In our setup, as in \cite{brockman2016openai}, each state $s$ is formed in the following way: 
\begin{equation*}
    s = [\theta, \dot{\theta}, \ddot{\theta}, goal - fk(\theta), goal]
\end{equation*}
Here, $\theta$ is the vector of joint angles of the simulated UR5 robot, $fk$
is a function maps joint angles $\theta$ to end effector position in Cartesian
space, and $goal$ is the current goal location in Cartesian space.  By including
the goal in the state description, we allow an agent to learn a policy
parameterized by the goal for a particular episode.  Our experimental setup
allows us to specify the number of joints controlled by a learning agent, so
$|s| = 3n + 6$, where $2 \leq n \leq 6$ is the number of joints being
controlled. The action vector $a$ is simply the vector of desired absolute
joint angles, and hence $|a| = n$. Finally, with a slight abuse of notation, we
formulate our tasks' reward functions as:
\begin{equation}
  \begin{split}
    \label{eq:reward}
    R(s, a) = - \Vert fk(s) &- s[goal] \Vert - \Vert a \Vert^2 + \\
      &100 \cdot \mathbb{I}\{fk(s) \in ball(s[goal], \epsilon)\}
  \end{split}
\end{equation}
This is a fairly standard sort of reward function definition.  The first term
in \eqref{eq:reward} penalizes the current Euclidean distance between
the agent's end effector and the goal position, the second term penalizes large
actions, and the final term gives a large positive reward when the agent
reaches the goal. This reward function is an example of a ``shaped'' reward,
which means that it attempts to steer the learning agent toward regions of high
reward. The first and second term in \eqref{eq:reward} accomplish this
by driving the agent to minimize the distance to the goal and to minimize the
sequence of controls necessary to accomplish this. Though one could argue that
in a position control regime it is inappropriate to penalize large absolute
actions, we do so here by analogy to velocity and torque control, in
which we would seek to minimize absolute actions to avoid sudden,
jerky motions. In contrast, a ``sparse'' reward would only provide the agent
with nonzero rewards upon reaching the goal (e.g., using just the third term in
\eqref{eq:reward} as a reward function).

\section{Results}
\label{sec:results}
We examine several general variants of the popular ``Reacher'' task,
exemplified in prior work by the \emph{Reacher-v2} (planar) and
\emph{FetchReach-v0} (3D) tasks in OpenAI Gym
\cite{brockman2016openai,todorov2012mujoco} \footnote{All results in this paper
were produced using the Docker container at
\url{https://hub.docker.com/r/cannon/testing}.}. We make use of a UR5 sitting
on an impermeable plane, which we simulate using ROS and our ROSGym interface.
Deep RL methods are commonly tested on discrete tasks, video games, and simple
control tasks. For these tasks, established simulation suites such as OpenAI
Gym and the Arcade Learning Environment \cite{bellemare2013arcade} are commonly
employed, but there is not currently a commonly accepted testing suite that
integrates well with existing control software for commercial robots.  This
limits the reproducibility and generality of results in robotic RL, as tasks
for new robots or tasks must often be hand-engineered.  It is our hope that,
when open-sourced, ROSGym will help researchers to validate results gathered
using existing robotic RL simulation suites (e.g.,
\cite{plappert2018multi,todorov2012mujoco}) on commercially available robots.

As in the established planar case, our experiments start UR5 robot
from a fixed position such that the end effector is within the goal
constraint region. We consider two broad categories of Reacher task: the
unconstrained version, in which goals are sampled from the whole workspace of
the robot, and several constrained versions, in which goals are sampled from a
goal constraint region.  For the unconstrained version of the task, the starting
joint angles are $[0,0,0,0,0,0]$. For the constrained versions, the
starting joint angles are $[0, \tfrac{\pi}{2}, \tfrac{\pi}{4}, 0, 0, 0]$. These
starting configurations are visualized in
Figures~\ref{fig:unconstrained}-\ref{fig:failconstraints}. The
learning agent is then tasked with controlling the end effector to a randomly
sampled goal location. We use joint position control as the action space of the
DDPG agent.  In general, our ROS--OpenAI Gym interface allows us to use joint
position, velocity, or torque control, but in our experiments only joint
position control resulted in a non-negligible success rate. It is also worth
noting that the choice of reward function had a significant impact on the
success rate achieved by DDPG on the Reacher tasks that we tested.

\def\arraystretch{1.2}
\begin{table}[hbt]
    \centering
    \caption{DDPG training hyperparameters}
    \label{table:hyperparams}
    \vspace*{-.5\baselineskip}
    \begin{tabular}{@{}c@{}c@{\hspace*{2pt}}c@{}}
        \toprule
        \textbf{Hyperparameter} & \textbf{Symbol} & \textbf{Value} \\
        \midrule
        Discount factor & $\gamma$ & 0.98 \\
        Replay Buffer Size & $B$ & $10^6$ \\
        Batch Size & $N$ & 64 \\
        Exploration Rate & $\nu$ & 0.01 \\
        Target Update Ratio & $\tau$ & 0.001 \\
        Actor Learning Rate & $\lambda_{\mu}$ & 0.0001 \\
        Critic Learning Rate & $\lambda_Q$ & 0.001 \\
        Episodes of Training & $M$ & 20,000 \\
        Steps per Episode & $T$ & 100 \\
    \end{tabular}
\end{table}

For all versions of the Reacher task described below, we derive goals by
uniformly sampling in the joint space of a simulated UR5 robot.  For each set
of joint values generated, we compute the end effector position by utilizing
the forward kinematics of the UR5. If the experiment involves a constraint
region, we reject candidate goals until a set of joint values places the
corresponding end effector location within the constraint region. Goals could
also be sampled directly in the Cartesian space within the goal constraint
region, but this could generate goal locations without
solutions for a given robot. Our goal sampling strategy
allows us to guarantee that each sampled goal location is hit by at least one
configuration of the robot. When we restrict to fewer than 6 joints
of the UR5, we sample goals within the reachable space effected by the joints
being controlled. 

The figures below show success rates for policies learned by DDPG over 20,000
episodes of training. Every 100 episodes of training, we conducted a testing
session comprising 100 episodes in which we execute the deterministic policy
learned by DDPG without exploration noise.  By default, episodes are 100 steps
long, which corresponds to 2 seconds of simulated time at a control rate of 50
Hz. Episodes are terminated early upon success, which we define as the robot's
end effector entering an $\epsilon$-ball surrounding the goal position, where
$\epsilon=0.1m$. The graphs below display the number of goals (out of 100) that
were successfully hit by the learned policy in each testing session. Each graph
shows the mean and 65\% confidence interval for 5 independent attempts at
learning using the same hyperparameters and task setup\footnote{Note that each
training run takes approximately 72 hours on an Amazon EC2 c4.xlarge instance.}.
We use the same actor and critic networks as \cite{lillicrap2015continuous}
uses for low-dimensional experiments.  Table~\ref{table:hyperparams} summarizes
the other hyperparameters for DDPG which are held constant across our
experiments.

In all of the following experiments, we strive to follow the prescriptions
presented by \cite{henderson2017deep}. We report all of our hyperparameters in
Table~\ref{table:hyperparams}, and report all runs of all experiments (i.e.,
the results shown in this paper were not cherry-picked to demonstrate the trend
that we highlight). Unfortunately, the baseline implementation of DDPG given by
OpenAI \cite{baselines} was not available when we conducted these experiments,
so we used our own codebase. However, we did replicate existing results using
our code before beginning the experiments reported in this work, and we intend
to open-source our code (including our implementation of DDPG and ROSGym).

\subsection{Unconstrained}

\begin{figure*}[hbt]
    \centering
    \subfigure[Joint numbers and bounding goal space region.]{\includegraphics[width=0.32\linewidth]{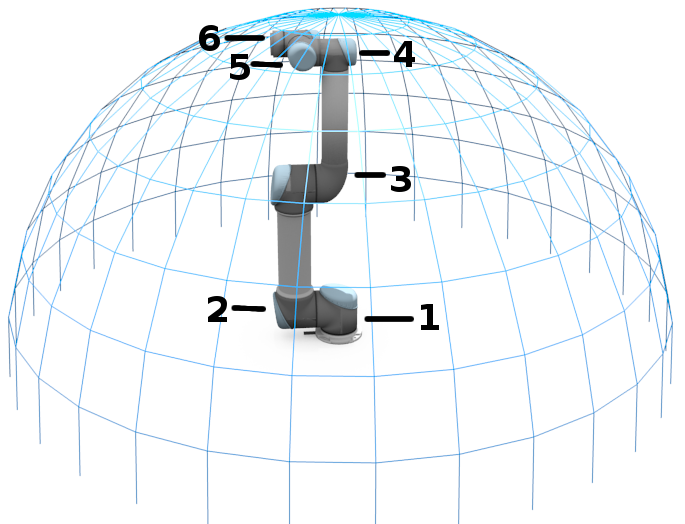}}
    \subfigure[Joints 1--2.]{\includegraphics[width=0.32\linewidth]{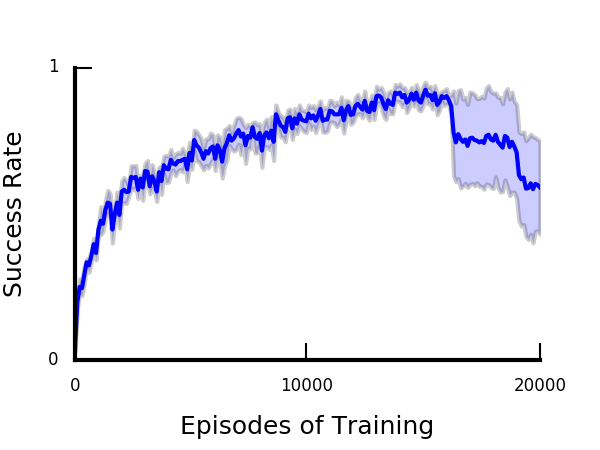}}
    \subfigure[Joints 1--3.]{\includegraphics[width=0.32\linewidth]{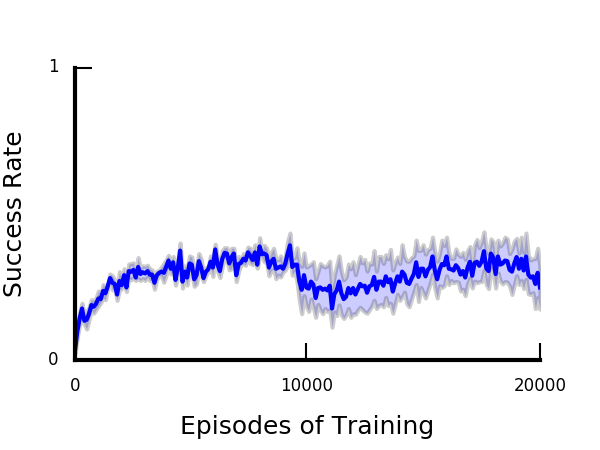}}

    \vspace*{-.5\baselineskip}\subfigure[Joints 1--4.]{\includegraphics[width=0.32\linewidth,trim=0 0 0 10,clip]{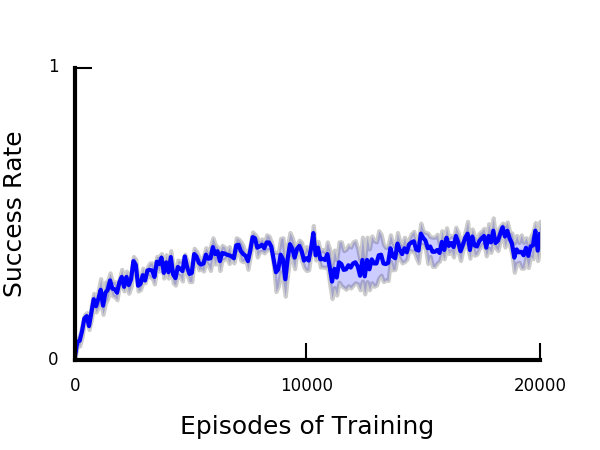}}
    \subfigure[Joints 1--5.]{\includegraphics[width=0.32\linewidth,trim=0 0 0 10,clip]{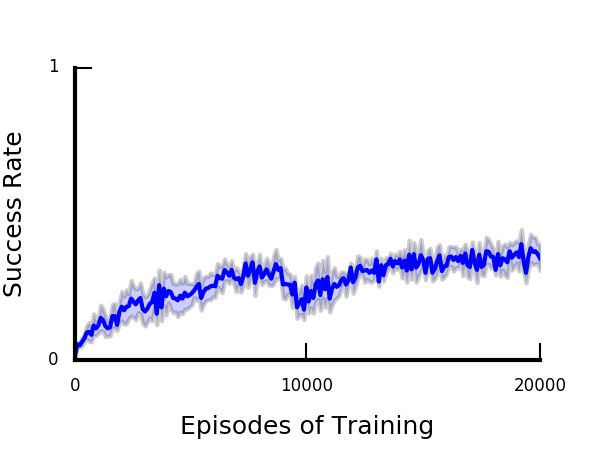}}
    \subfigure[Joints 1--6.]{\includegraphics[width=0.32\linewidth,trim=0 0 0 10,clip]{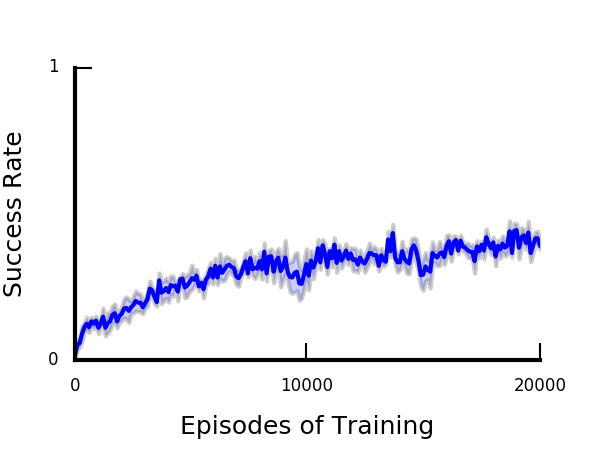}}
    \vspace*{-.5\baselineskip}\caption{Results for the unconstrained case.}
    \label{fig:unconstrained}
\end{figure*}

Figure~\ref{fig:unconstrained} shows the results of running the variant of DDPG
described in Section~\ref{sec:methods} on the unconstrained Reacher task. In
this version of the task we place no constraint on the sampling of goal end
effector locations; hence, goals are sampled throughout the robot's workspace.
It is easy to see that, while DDPG succeeds in learning a highly capable policy
for 2 joint control, for 3--6 joints DDPG fails to learn a policy that can hit
more than 50\% of the sampled goal locations. This makes some intuitive sense,
as the 2 joints at the base of the UR5 effect roughly orthogonal motions and
only a simple control policy must be learned, whereas for 3--6 joints a
competent policy must involve coupled motion of multiple joints.

It is interesting to note that the asymptotic behavior of DDPG is approximately
equivalent for 3--6 joints, as shown in Figure~\ref{fig:unconstrained}. This
may not be particularly surprising, however; we note that the Cartesian
workspaces for 3, 4, 5, and 6 joints of the UR5 robot are almost identical,
since joints 4, 5, and 6 are primarily responsible for the orientation, rather
than the position, of the end effector. For the unconstrained experiment we
reported training curves for all of these joint configurations in order to
verify that learning occurs similarly for all joint variants, but for all
following experiments we report 3 and 6 joint results as these provide a lower
and upper bound on learning difficulty, respectively\footnote{Apart from the
2-joint variant, which was learned easily in the unconstrained case and in all
other cases, and so is not further analyzed in this paper.}.

\subsection{Z-Height and Close Box Constraints}

\begin{figure*}[bt]
    \centering
    \subfigure[Z-height goal constraint.]{\label{subfig:zheightviz} \includegraphics[width=0.32\linewidth]{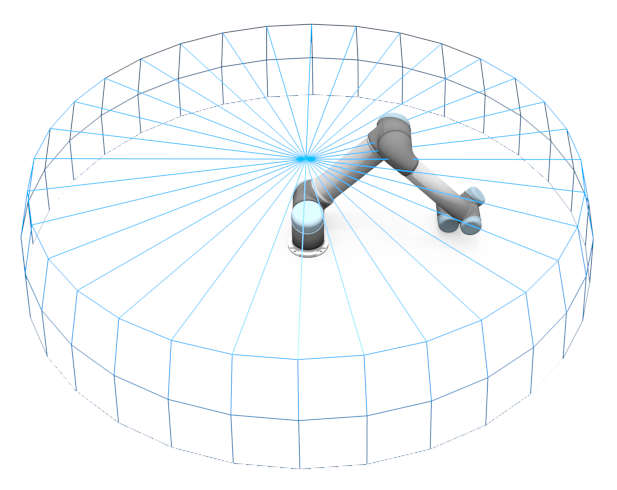}}
    \subfigure[Joints 1--3, z-height.]{\includegraphics[width=0.32\linewidth]{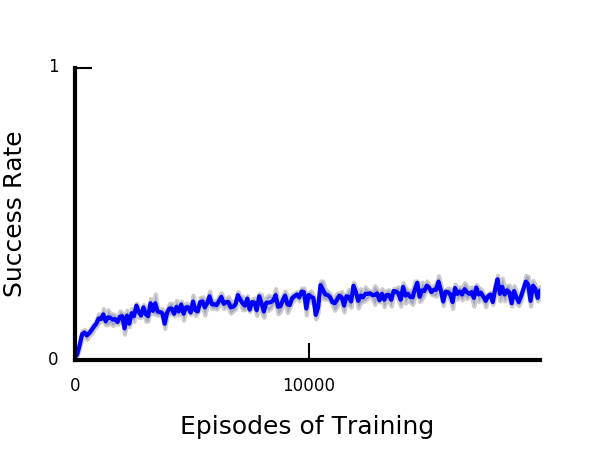}}
    \subfigure[Joints 1--6, z-height.]{\includegraphics[width=0.32\linewidth]{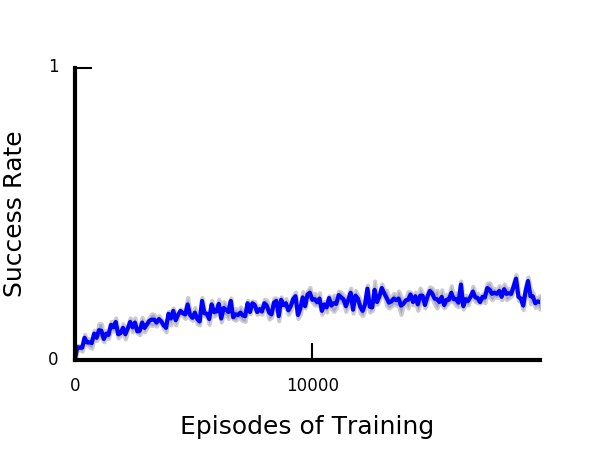}}

    \vspace*{-\baselineskip}\subfigure[Close box goal constraint.]{\label{subfig:closeboxviz} \includegraphics[width=0.32\linewidth]{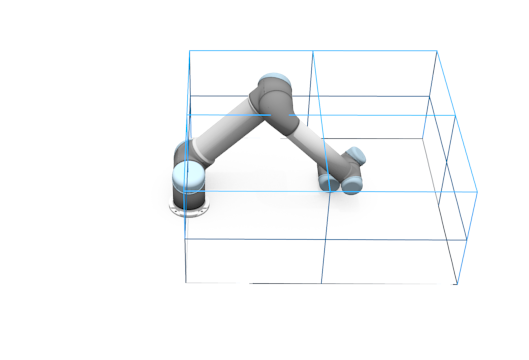}}
    \subfigure[Joints 1--3, close box.]{\includegraphics[width=0.32\linewidth]{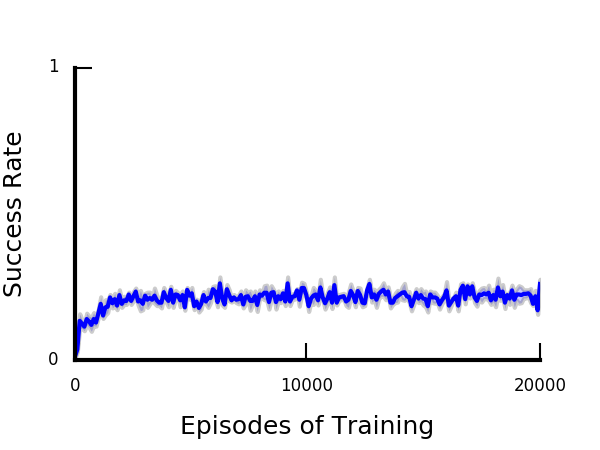}}
    \subfigure[Joints 1--6, close box.]{\includegraphics[width=0.32\linewidth]{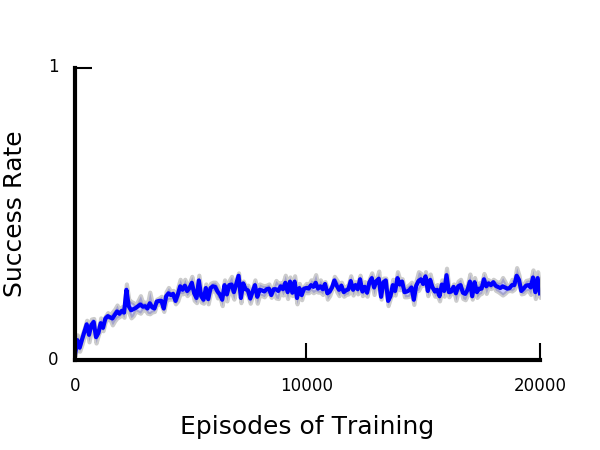}}

    \vspace*{-\baselineskip}\subfigure[Far box goal constraint.]{\label{subfig:farboxviz} \includegraphics[width=0.32\linewidth]{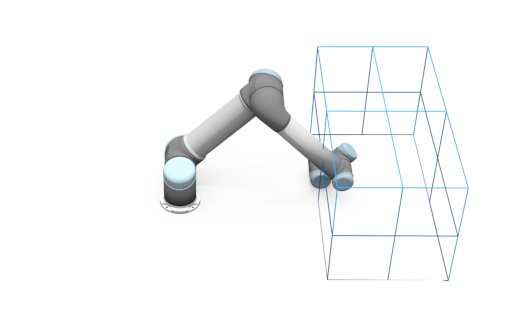}}
    \subfigure[Joints 1--3, far box.]{\includegraphics[width=0.32\linewidth]{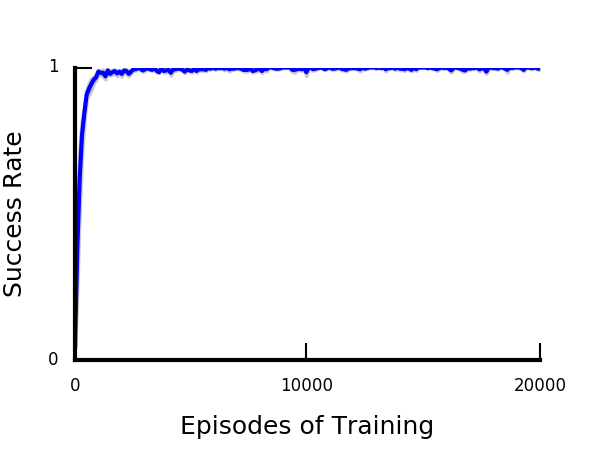}}
    \subfigure[Joints 1--6, far box.]{\label{subfig:finalviz}\includegraphics[width=0.32\linewidth]{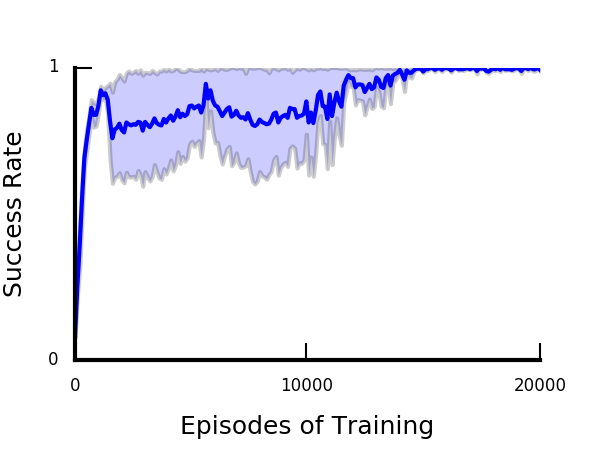}}

    \caption{Z-height and close box constrained experiments.}
    \vspace*{-\baselineskip}\label{fig:failconstraints}
\end{figure*}

Figure~\ref{fig:failconstraints} shows the results of running DDPG on the
z-height constrained and close box constrained versions of the Reacher task. In
the z-height constrained version, goals are only sampled below a height of 0.4
meters from the plane on which the robot sits. Note that the end effector of
the simulated UR5 can normally reach approximately 1.0 meters above the plane
on which the robot sits. In the close box constrained version, goals are sampled
from a $0.9m \times 0.8m \times 0.4m$ box that includes the robot's base. See
Figure~\ref{subfig:zheightviz} and Figure~\ref{subfig:closeboxviz} for
visualizations of these goal constraint regions. We only show the 3 and 6 joint
cases here, as DDPG is able to learn the unconstrained task well for 2 joint
control, and throughout our experiments we saw little variation in learning
among the 3, 4, 5, and 6 joint versions of this task. 

Though these two forms of constraint are quite different workspace regions, the
asymptotic performance of DDPG on these constraint regions is identical. Note that
the asymptotic success rate of the policy learned by DDPG decreases from
approximately 40\% in the unconstrained case to 20\% in the z-height and close
box constrained cases. This lends further evidence to the idea that certain
regions of the robot's workspace (such as the region above z = $0.4$m) are easier
to learn than others (such as the constraint regions previously described).
Finally, we can qualitatively observe that learning takes longer to converge
for 6 joint control, which lends some validation to the conventional wisdom
that RL scales poorly to higher dimensions. However, this scaling effect is
strongly dominated by the similarly poor asymptotic behavior common to the 3
and 6 joint cases. 

\subsection{Far Box Constraint}

%

Figures~\ref{subfig:farboxviz} and \ref{subfig:finalviz} show the results of running DDPG on the far box
constrained version of the Reacher task. In this version, goals are only
sampled from within a $0.4m \times 0.8m \times 0.4m$ box which is $0.5$ meters
along the $X$-axis from the robot's base. The dimensions of this box were changed
because the robot's workspace does not extend beyond $0.9$ meters from the base
of the robot.  Though the volume of this constraint region is approximately
half of that of the goal constraint region in the close box case, in our
experiments an extension of the height of the far box constraint region to
$1.0$ meters resulted in a convergent success rate similar to that in the
reported far box case. See Figure~\ref{subfig:farboxviz} for a visualization of
the reported far box constraint region. As before, we only show the 3 and 6
joint cases here. 

It is immediately apparent that the far box constraint is easier to learn than
any of the previously considered goal regions. Learning converges to nearly
100\% success within 10 testing sessions, or 1000 training episodes, for all
independent runs in the 3 joint case, and for 4 of 5 runs in the 6 joint case
(1 run experienced the type of catastrophic forgetting that DDPG is known
for \cite{duan2016benchmarking}).
We also found this rapid learning and near-perfect asymptotic behavior holds
even if the z-height constraint of the box is removed. By simply moving the
goal sampling region away from the base of the robot, the apparent asymptotic
performance and sample complexity of DDPG on this task are significantly
improved. Note that this form of constraint is the most similar to the goal
constraints in existing Reacher task implementations, and also displays the
greatest ease of learning.

\begin{figure*}[bt]
    \centering
    \vspace*{-\baselineskip}
    \subfigure[Run 1, 0.7 $\leq z \leq 0.8$.]{\includegraphics[width=0.45\linewidth,trim=0 10 0 0,clip]{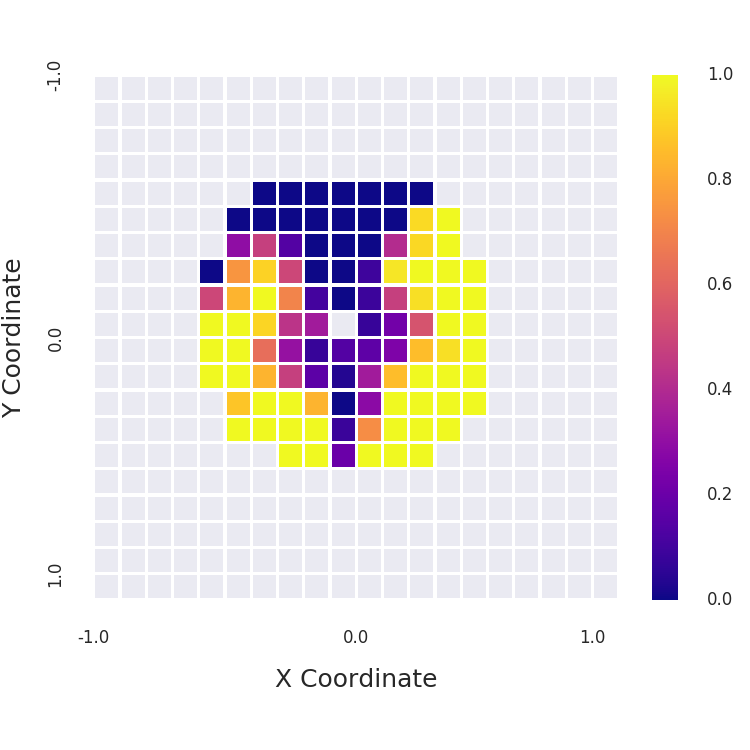}}
    \subfigure[Run 2, 0.7 $\leq z \leq 0.8$.]{\includegraphics[width=0.45\linewidth,trim=0 10 0 0,clip]{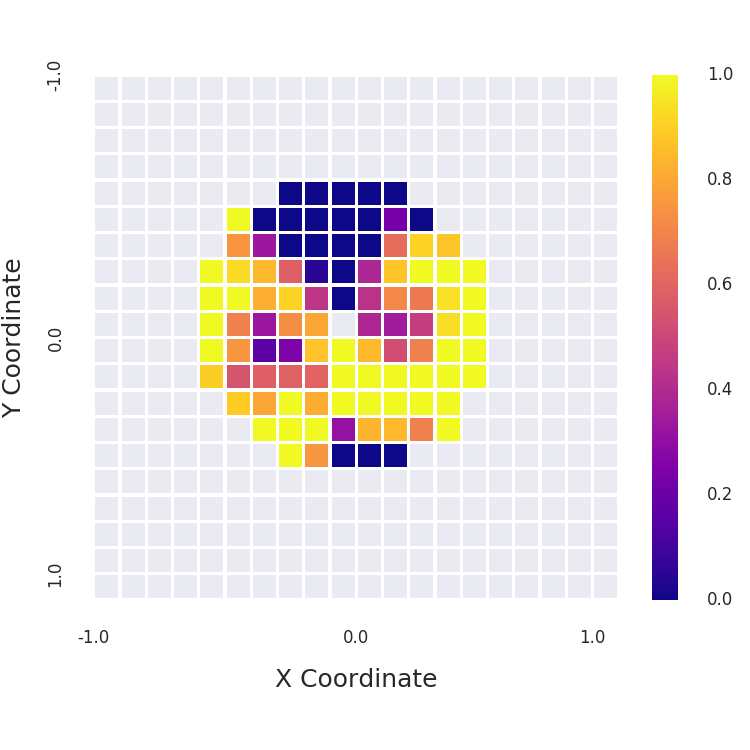}}
    \vspace*{-0.5\baselineskip}\caption{Goal success regions for independent runs on unconstrained 3 joint
    task.}
    \label{fig:goal_success}
\end{figure*}

\subsection{Policy Success Visualization}

Finally, Figure~\ref{fig:goal_success} shows how two runs of DDPG can learn two
very different policies. These visualizations were created by taking two
convergent ($\approx40\%$ success) policies learned by DDPG as described for the
unconstrained 3-joint control Reacher task and running them without exploration
noise. We measured the two policies' rates of success on a coarse tiling of the
workspace of the UR5 by randomly sampling $10,000$ goal locations in the
previously described manner, executing the learnt policy, and binning the successes for each grid cell. The figures above represent only a
slice of the robot's workspace (between $z = 0.7$ and $z = 0.8$). 

Though they were learned on the same task and with the same hyperparameters, the
two policies display very distinct regions of competence (lighter regions,
where nearly 100\% of goals are successfully hit). This could result from DDPG
attempting to learn a single unifying policy for regions of differing
difficulty. From the qualitative difference in these two policies, we can infer
that biases at the start of training may have an inordinate effect on the
resulting policy success regions.  This phenomenon could account for recent
successes in robotic RL: tasks have been restricted to small enough spaces that
a single policy can be learned which accounts for the entire constrained space.

\section{Discussion}
\label{sec:discussion}
Our results strongly suggest that 1) more exploration is necessary into how
Deep Robotic RL benchmarks should be defined and run and that 2) more work is
needed before popular Deep RL methods will be capable of learning control
policies for general robotic tasks. Standard robotic RL benchmark tasks can
elide much of what makes robotic control difficult, and much is still unknown
about the true capability of popular algorithms such as DDPG to learn to
perform more general tasks. Recent work has established that Deep RL methods
may be more difficult to generalize to complex tasks than prior, non-deep
methods \cite{NIPS2017_7233,mania2018simple,mahmood2018setting}, but it is
still unknown how broadly this applies. However, this does not necessarily mean
that the more general forms of these tasks are impossible to learn; rather, it
suggests that more work needs to be done in assessing how popular learning
algorithms perform on non-constrained versions of robotic tasks. Given the
efficacy of ensemble methods \cite{dietterich2000ensemble} for supervised
learning, we expect that a regional Deep RL ensemble algorithm may perform
better on the Reacher tasks considered in this paper. We fixed our choice of
algorithm and hyperparameters in this paper to focus on the effect of varying
the ``Reacher'' task definition on learning, but we have not yet attempted the
same analysis with other popular Deep RL algorithms such as TRPO.  We expect,
however, to observe a similar trend of learning difficulty as the goal
constraint region is expanded and the dynamics of the robot's effective
workspace become more complicated. More research is necessary to confirm that
the results presented in this paper are general across Deep RL algorithms, but
our initial analysis has produced unexpected learning behavior that merits
further investigation.

\section*{Acknowledgments}
We would like to thank Zak Kingston and Bryce Willey for their help in
developing ROSGym and this manuscript.

\FloatBarrier

\IEEEtriggeratref{22}
\bibliographystyle{IEEEtran}
\bibliography{bib/main}

\begin{thebibliography}{10}
\providecommand{\url}[1]{#1}
\csname url@samestyle\endcsname
\providecommand{\newblock}{\relax}
\providecommand{\bibinfo}[2]{#2}
\providecommand{\BIBentrySTDinterwordspacing}{\spaceskip=0pt\relax}
\providecommand{\BIBentryALTinterwordstretchfactor}{4}
\providecommand{\BIBentryALTinterwordspacing}{\spaceskip=\fontdimen2\font plus
\BIBentryALTinterwordstretchfactor\fontdimen3\font minus
  \fontdimen4\font\relax}
\providecommand{\BIBforeignlanguage}[2]{{%
\expandafter\ifx\csname l@#1\endcsname\relax
\typeout{** WARNING: IEEEtran.bst: No hyphenation pattern has been}%
\typeout{** loaded for the language `#1'. Using the pattern for}%
\typeout{** the default language instead.}%
\else
\language=\csname l@#1\endcsname
\fi
#2}}
\providecommand{\BIBdecl}{\relax}
\BIBdecl

\bibitem{sutton1998reinforcement}
R.~S. Sutton and A.~G. Barto, \emph{Reinforcement learning: An
  introduction}.\hskip 1em plus 0.5em minus 0.4em\relax MIT press Cambridge,
  1998.

\bibitem{gu2017deep}
S.~Gu, E.~Holly, T.~Lillicrap, and S.~Levine, ``Deep reinforcement learning for
  robotic manipulation with asynchronous off-policy updates,'' in \emph{IEEE
  International Conference on Robotics and Automation}, Oct. 2017, pp.
  3389--3396.

\bibitem{levine2016end}
S.~Levine, C.~Finn, T.~Darrell, and P.~Abbeel, ``End-to-end training of deep
  visuomotor policies,'' \emph{Journal of Machine Learning Research}, vol.~17,
  no.~1, pp. 1334--1373, 2016.

\bibitem{vevcerik2017leveraging}
M.~Ve{\v{c}}er{\'{i}}k, T.~Hester, J.~Scholz, F.~Wang, O.~Pietquin, B.~Piot,
  N.~Heess, T.~Roth{\"{o}}rl, T.~Lampe, and M.~Riedmiller, ``Leveraging
  demonstrations for deep reinforcement learning on robotics problems with
  sparse rewards,'' \emph{arXiv preprint arXiv:1707.08817}, Jul. 2017.

\bibitem{andrychowicz2017hindsight}
M.~Andrychowicz, F.~Wolski, A.~Ray, J.~Schneider, R.~Fong, P.~Welinder,
  B.~McGrew, J.~Tobin, P.~Abbeel, and W.~Zaremba, ``Hindsight experience
  replay,'' in \emph{Advances in Neural Information Processing Systems}, 2017,
  pp. 5048--5058.

\bibitem{duan2016benchmarking}
Y.~Duan, X.~Chen, R.~Houthooft, J.~Schulman, and P.~Abbeel, ``Benchmarking deep
  reinforcement learning for continuous control,'' in \emph{International
  Conference on Machine Learning}, 2016, pp. 1329--1338.

\bibitem{henderson2017deep}
\BIBentryALTinterwordspacing
P.~Henderson, R.~Islam, P.~Bachman, J.~Pineau, D.~Precup, and D.~Meger, ``Deep
  reinforcement learning that matters,'' in \emph{AAAI Conference on Artificial
  Intelligence}, 2018, pp. 3207--3214. [Online]. Available:
  \url{https://www.aaai.org/ocs/index.php/AAAI/AAAI18/paper/view/16669}
\BIBentrySTDinterwordspacing

\bibitem{brockman2016openai}
G.~Brockman, V.~Cheung, L.~Pettersson, J.~Schneider, J.~Schulman, J.~Tang, and
  W.~Zaremba, ``{OpenAI Gym},'' \emph{arXiv preprint arXiv:1606.01540}, 2016.

\bibitem{todorov2012mujoco}
E.~Todorov, T.~Erez, and Y.~Tassa, ``{MuJoCo}: A physics engine for model-based
  control,'' in \emph{IEEE International Conference on Intelligent Robots and
  Systems}, 2012, pp. 5026--5033.

\bibitem{plappert2018multi}
M.~Plappert, M.~Andrychowicz, A.~Ray, B.~McGrew, B.~Baker, G.~Powell,
  J.~Schneider, J.~Tobin, M.~Chociej, P.~Welinder, V.~Kumar, and W.~Zaremba,
  ``Multi-goal reinforcement learning: Challenging robotics environments and
  request for research,'' \emph{arXiv preprint arXiv:1802.09464}, 2018.

\bibitem{Kober2013}
J.~Kober, J.~A. Bagnell, and J.~Peters, ``Reinforcement learning in robotics: A
  survey,'' \emph{International Journal of Robotics Research}, vol.~32, no.~11,
  pp. 1238--1278, 2013.

\bibitem{kohl2004policy}
N.~Kohl and P.~Stone, ``Policy gradient reinforcement learning for fast
  quadrupedal locomotion,'' in \emph{IEEE International Conference on Robotics
  and Automation}, 2004, pp. 2619--2624.

\bibitem{craig2005introduction}
J.~J. Craig, \emph{Introduction to robotics: mechanics and control},
  3rd~ed.\hskip 1em plus 0.5em minus 0.4em\relax Upper Saddle River, NJ, USA:
  Pearson/Prentice Hall, 2005.

\bibitem{lillicrap2015continuous}
T.~P. Lillicrap, J.~J. Hunt, A.~Pritzel, N.~Heess, T.~Erez, Y.~Tassa,
  D.~Silver, and D.~Wierstra, ``Continuous control with deep reinforcement
  learning,'' in \emph{International Conference on Learning Representations},
  Sep. 2016.

\bibitem{quigley2009ros}
M.~Quigley, K.~Conley, B.~Gerkey, J.~Faust, T.~Foote, J.~Leibs, R.~Wheeler, and
  A.~Y. Ng, ``{ROS}: an open-source {R}obot {O}perating {S}ystem,'' in
  \emph{ICRA workshop on open source software}, 2009.

\bibitem{mnih2015human}
V.~Mnih, K.~Kavukcuoglu, D.~Silver, A.~A. Rusu, J.~Veness, M.~G. Bellemare,
  A.~Graves, M.~Riedmiller, A.~K. Fidjeland, G.~Ostrovski \emph{et~al.},
  ``Human-level control through deep reinforcement learning,'' \emph{Nature},
  vol. 518, no. 7540, p. 529, 2015.

\bibitem{schulman2015trust}
J.~Schulman, S.~Levine, P.~Abbeel, M.~Jordan, and P.~Moritz, ``Trust region
  policy optimization,'' in \emph{International Conference on Machine
  Learning}, 2015, pp. 1889--1897.

\bibitem{haarnoja2017reinforcement}
T.~Haarnoja, H.~Tang, P.~Abbeel, and S.~Levine, ``Reinforcement learning with
  deep energy-based policies,'' \emph{arXiv preprint arXiv:1702.08165}, 2017.

\bibitem{peng2018deepmimic}
X.~B. Peng, P.~Abbeel, S.~Levine, and M.~van~de Panne, ``{DeepMimic}:
  Example-guided deep reinforcement learning of physics-based character
  skills,'' \emph{arXiv preprint arXiv:1804.02717}, 2018.

\bibitem{gu2016continuous}
S.~Gu, T.~Lillicrap, I.~Sutskever, and S.~Levine, ``Continuous deep
  {Q}-learning with model-based acceleration,'' in \emph{International
  Conference on Machine Learning}, Mar. 2016, pp. 2829--2838.

\bibitem{rusu2016sim}
A.~A. Rusu, M.~Vecerik, T.~Roth{\"o}rl, N.~Heess, R.~Pascanu, and R.~Hadsell,
  ``Sim-to-real robot learning from pixels with progressive nets,'' \emph{arXiv
  preprint arXiv:1610.04286}, 2016.

\bibitem{NIPS2017_7233}
A.~Rajeswaran, K.~Lowrey, E.~V. Todorov, and S.~M. Kakade, ``Towards
  generalization and simplicity in continuous control,'' in \emph{Advances in
  Neural Information Processing Systems 30}, I.~Guyon, U.~V. Luxburg,
  S.~Bengio, H.~Wallach, R.~Fergus, S.~Vishwanathan, and R.~Garnett, Eds.\hskip
  1em plus 0.5em minus 0.4em\relax Curran Associates, Inc., 2017, pp.
  6550--6561.

\bibitem{schaul2015prioritized}
T.~Schaul, J.~Quan, I.~Antonoglou, and D.~Silver, ``Prioritized experience
  replay,'' \emph{arXiv preprint arXiv:1511.05952}, 2015.

\bibitem{bellemare2013arcade}
M.~G. Bellemare, Y.~Naddaf, J.~Veness, and M.~Bowling, ``The arcade learning
  environment: An evaluation platform for general agents,'' \emph{Journal of
  Artificial Intelligence Research}, vol.~47, pp. 253--279, 2013.

\bibitem{baselines}
P.~Dhariwal, C.~Hesse, O.~Klimov, A.~Nichol, M.~Plappert, A.~Radford,
  J.~Schulman, S.~Sidor, Y.~Wu, and P.~Zhokhov, ``Openai baselines,''
  \url{https://github.com/openai/baselines}, 2017.

\bibitem{mania2018simple}
H.~Mania, A.~Guy, and B.~Recht, ``Simple random search provides a competitive
  approach to reinforcement learning,'' \emph{arXiv preprint arXiv:1803.07055},
  2018.

\bibitem{mahmood2018setting}
A.~R. Mahmood, D.~Korenkevych, B.~J. Komer, and J.~Bergstra, ``Setting up a
  reinforcement learning task with a real-world robot,'' \emph{arXiv preprint
  arXiv:1803.07067}, 2018.

\bibitem{dietterich2000ensemble}
T.~G. Dietterich, ``Ensemble methods in machine learning,'' in
  \emph{International workshop on multiple classifier systems}.\hskip 1em plus
  0.5em minus 0.4em\relax Springer, 2000, pp. 1--15.

\end{thebibliography}

\end{document}